\documentclass{article}
\usepackage{iclr2024_conference,times}
\usepackage{graphicx}
\usepackage{makecell}
\usepackage{booktabs}
\usepackage{multirow}
\usepackage{amssymb}
\usepackage{wrapfig}
\usepackage{float}

\usepackage{amsmath,amsfonts,bm}

\def\eqref#1{equation~\ref{#1}}

\def\1{\bm{1}}

\DeclareMathAlphabet{\mathsfit}{\encodingdefault}{\sfdefault}{m}{sl}
\SetMathAlphabet{\mathsfit}{bold}{\encodingdefault}{\sfdefault}{bx}{n}

\def\sI{{\mathbb{I}}}

\usepackage{hyperref}
\usepackage{url}

\makeatletter
\newcommand{\rmnum}[1]{\romannumeral #1}
\newcommand{\Rmnum}[1]{\expandafter\@slowromancap\romannumeral #1@}
\makeatother
\graphicspath{{figs/}}

\title{Investigating the Impact of Quantization on Adversarial Robustness}

\vspace{-0.2cm}
\author{Qun Li, Yuan Meng\footnotemark[2]\hspace{0.5em}, Chen Tang, Jiacheng Jiang, Zhi Wang\footnotemark[2]\\
Tsinghua University\\
\texttt{liq23@mails.tsinghua.edu.cn, yuanmeng@mail.tsinghua.edu.cn}\\ \texttt{genprtung@gmail.com, jiangjc23@mails.tsinghua.edu.cn}\\
\texttt{wangzhi@sz.tsinghua.edu.cn}
}

\iclrfinalcopy
\begin{document}

\renewcommand{\thefootnote}{\fnsymbol{footnote}}
\footnotetext[2]{Corresponding authors.}

\maketitle

\vspace{-0.2cm}

\begin{abstract}

\vspace{-0.2cm}
Quantization is a promising technique for reducing the bit-width of deep models to improve their runtime performance and storage efficiency, and thus becomes a fundamental step for deployment. In real-world scenarios, quantized models are often faced with adversarial attacks which cause the model to make incorrect inferences by introducing slight perturbations. However, recent studies have paid less attention to the impact of quantization on the model robustness. More surprisingly, existing studies on this topic even present inconsistent conclusions, which prompted our in-depth investigation. In this paper, we conduct a first-time analysis of the impact of the quantization pipeline components that can incorporate robust optimization under the settings of Post-Training Quantization and Quantization-Aware Training. Through our detailed analysis, we discovered that this inconsistency arises from the use of different pipelines in different studies, specifically regarding whether robust optimization is performed and at which quantization stage it occurs. Our research findings contribute insights into deploying more secure and robust quantized networks, assisting practitioners in reference for scenarios with high-security requirements and limited resources.

\end{abstract}

\vspace{-0.2cm}
\section{Introduction}

Deep neural networks have demonstrated outstanding performance in tasks such as computer vision (\citealp{ronneberger2015unet, howard2017mobilenets}) and natural language processing (\citealp{sutskever2014sequence, vaswani2017attention}). However, the increasing number of model parameters makes it challenging to deploy them in resource-constrained scenarios. 
As a model compression technique, quantization effectively reduces the inference memory and time cost by quantizing model parameters and/or activations from full-precision to low-bit integers (\citealp{whitepaper2021}). Current research aims to achieve better performance with lower bit-width (\citealp{pact, esser2019lsq}). Many quantization methods (\citealp{yao2022zeroquant}) have been incorporated into the deep learning libraris (\citealp{rasley2020deepspeed}). However, many resource-constrained scenarios require not only efficiency but also high security for model inference, such as in autonomous driving (\citealp{liu2021auto-drive1, katare2023auto-drive2}) and medical assistance (\citealp{zhang2021medical1}). These applications often face the risk of adversarial attacks, where imperceptible perturbations are added to input samples, leading the model to make incorrect predictions (\citealp{pgd}). This poses a fatal risk in certain scenarios. Due to adversarial robustness being at odds with accuracy (\citealp{pmlr-v202-hu23j}), it indicates that considering both quantization and adversarial robustness simultaneously is necessary.

\begin{wrapfigure}[8]{r}{0.4\textwidth}

\vspace{-0.6cm}
\begin{center}
\includegraphics[width=0.4\textwidth]{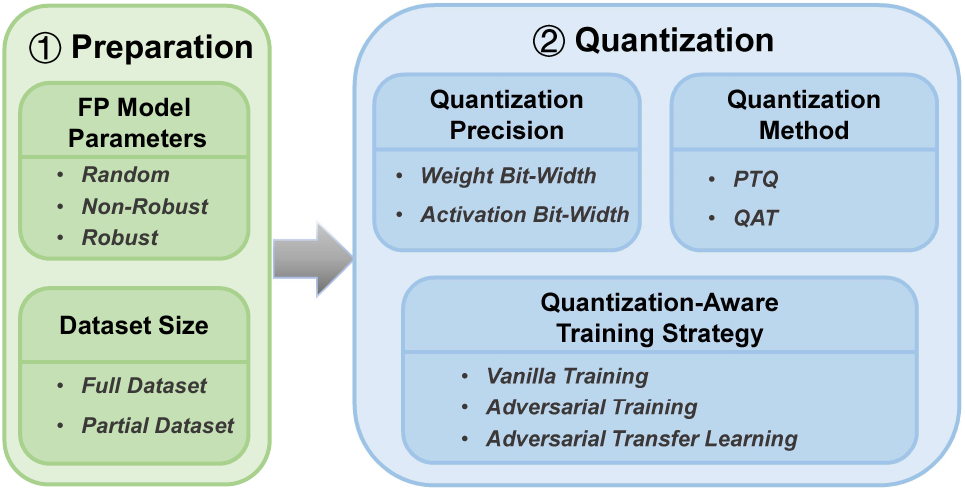}
\end{center}
\vspace{-0.5cm}
\caption{Quantization Pipeline. }
\label{pipeline}
\end{wrapfigure}

Existing research has focused on improving the adversarial robustness of quantized networks. However, we find that the conclusions drawn from these studies are conflicting. For instance, some studies observe that as the quantization bit-width decreases, adversarial robustness deteriorates (\citealp{lin2021integer, sen2020empir, xiao2023robustmq}), while others reach the opposite conclusion (\citealp{fu2021double, lin2019defensive}). We find the conflicting results arise from differences in the pipeline settings during the quantization process and the possible addition of robustness components at different stages.

To address this question, we first clarify the pipeline of the quantization process, as shown in Figure \ref{pipeline}. This involves the \textbf{preparation stage} and the \textbf{quantization stage}. In the preparation stage, it is necessary to specify whether a pre-trained full-precision model (FP model) is employed and whether it undergoes adversarial training, which lead to three options: \textit{robust parameters, non-robust parameters, and random parameters}. Additionally, determining the size of the dataset for calibration or fine-tuning is essential. In the quantization stage, the choice between Quantization-Aware Training (QAT) and Post-Training Quantization (PTQ) needs to be made. It is also needed to determine whether weights and activations need quantization, along with specifying the quantization bit-width. For QAT, whether adversarial training is incorporated into the process is also needed to ascertain.
Please refer to Appendix A for the details of Quantization-Aware Training Strategy.
A variety of choices make it difficult for researchers to determine the pipeline, and it is uncertain about the impact of each choice on robustness.

\begin{table}[t]
\caption{The impact of different quantization pipelines on robustness. (\textcolor{red}{\large \textbf +}) means reducing bit-width enhances robustness, while (\textcolor[RGB]{84, 179, 69}{\large \textbf {--}}) indicates the opposite, and ({\small \textcolor{blue}{$\boldsymbol{\sim}$}}) means moderate bit-width gains the best robustness. $\epsilon$ means the adversarial attack intensity (\citealp{pgd}), and we define $\epsilon \leq 2$ as the weaker attack, and $\epsilon \leq 8$ as the normal attack.
Additional details regarding attack intensity and quantization settings will be supplemented in Appendix A.} 
\vspace{-0.3cm}
\label{impact-different-pipelines}
\begin{center}
\resizebox{\textwidth}{!}{
\renewcommand{\arraystretch}{0.5}
\begin{tabular}{llll}
\toprule[1.6pt] \\
\multicolumn{1}{c}{\bf Quantization Setting}   
&\multicolumn{1}{c}{\bf Random Parameters}
&\multicolumn{1}{c}{\bf Non-Robust Parameters} 
&\multicolumn{1}{c}{\bf Robust Parameters}  
\\ \\ \midrule \\
    \makecell[c]{PTQ}
    &{\makecell[c]{------}}
    &{\makecell[c]{Against weaker attack, lower\\bit-width leads to higher robustness (\textcolor{red}{\large \textbf +}).}}
    &{\makecell[c]{Against normal attack, lower \\ bit-width leads to lower robustness (\textcolor[RGB]{84, 179, 69}{\large \textbf {--}}).}}
\\ \\ \midrule \\
    \makecell[c]{QAT\\(Vanilla Training)}
    &{\makecell[c]{Against weaker attack, lower \\bit-width leads to higher robustness (\textcolor{red}{\large \textbf +}).}}
    &{\makecell[c]{Against weaker attack, lower \\bit-width leads to higher robustness (\textcolor{red}{\large \textbf +}).}}
    &{\makecell[c]{Against weaker attack, lower \\bit-width leads to higher robustness (\textcolor{red}{\large \textbf +}).}}
\\ \\\midrule \\
    \makecell[c]{QAT\\(Adversarial Training)}       
    &{\makecell[c]{Against normal attack, lower\\ bit-width leads to lower robustness (\textcolor[RGB]{84, 179, 69}{ \large \textbf {--}}).}}
    &{\makecell[c]{Accuracy significantly decreases compared to \\initialization, against normal attack,\\ lower bit-width leads to higher robustness (\textcolor{red}{\large \textbf +}).}}
    &{\makecell[c]{Against normal attack, moderate \\bit-width exhibits the best robustness ({\small \textcolor{blue}{$\boldsymbol{\sim}$}}).}}
\\ \\ \midrule \\
    \makecell[c]{QAT\\(Adversarial Transfer Learning)} 
    &{\makecell[c]{Against normal attack, but less robustness\\than robust initialized ways, lower\\bit-width leads to lower robustness ({\large \textcolor[RGB]{84, 179, 69}{\textbf {--}}}).}}
    &{\makecell[c]{Against normal attack, but less robustness\\than robust initialized ways, moderate \\bit-width exhibits the best robustness ({\small \textcolor{blue}{$\boldsymbol{\sim}$}}).}}
    &{\makecell[c]{Against normal attack, lower \\bit-width leads to lower robustness ({\large \textcolor[RGB]{84, 179, 69}{\textbf {--}}}).}}
\\ \\ \bottomrule[1.6pt]
\end{tabular}
}
\vspace{-0.5cm}
\end{center}
\end{table}

We experiment with different pipelines to observe their robustness. The key conclusions include: (\rmnum{1}) Quantization without any robustness components demonstrates resistance to lower levels of adversarial attacks, and a lower bit-width for quantization leads to improved robustness. (\rmnum{2}) In certain pipelines, the quantized model can achieve robustness similar to that of the full precision robust model. In such cases, a lower quantization bit-width may result in decreased robustness. (\rmnum{3}) Adding adversarial training to quantization can gain robustness, however, using the wildly used PGD-7 adversarial training incurs an additional $7\times$ time overhead. Indeed, we have analyzed all possible scenarios, and the conclusions are discussed in Table \ref{impact-different-pipelines}.

\vspace{-0.15cm}
\section{Related Work}
\vspace{-0.15cm}
As early as 2017, \cite{galloway2017attacking} reveals that the binarized models exhibit adversarial robustness, but it is later known this comes from obfuscated gradients (\citealp{obfuscated-gradients}). 
Subsequently, \cite{lin2019defensive} argues that quantized networks suffer from error amplification effects, making them more vulnerable to adversarial attacks. They address this risk by designing regularization terms. 
\cite{sen2020empir} claims that there is an accuracy-robustness trade-off during quantization, and assembles full-precision and quantized networks by ensembling methods to better reconcile this trade-off. 
\cite{fu2021double} further enhances model robustness by designing a stochastic precision inference scheme. 
However, these studies are hard to reveal the real tendency of quantization robustness, as they are not conducted at a unified pipeline, which even leads to conflicting conclusions.

Another parallel set of works involves evaluating the robustness and reliability of quantized networks. 
\cite{xiao2023robustmq} only assesses the robustness of QAT without any robust components, while \cite{yuan2023reliability-benchmarking} discusses aspects such as category performance differences and out-of-distribution performance instead of adversarial robustness on Post-Training Quantization networks. In contrast, our paper focuses on the adversarial robustness of PTQ and QAT, discussing the impact of various choices on the components of the quantization pipeline.

\section{Robust Quantization Overview}

In this section, we first introduce the pipeline for model quantization, extracting components that may influence the adversarial robustness of the model in the process. These components will be analyzed in the next section. Furthermore, we highlight the evaluation of quantized models with adversarial robustness metrics in resource-constrained environments.

\subsection{Quantization Pipeline}

As shown in Figure \ref{pipeline}, whether it is  PTQ or QAT, we can divide the quantization process into two key stages: the preparation stage and the quantization stage. First, it is necessary to determine the initialization parameters of the model and the size of the dataset used for calibration or fine-tuning. Then, the bit-width for quantization and the training strategy for fine-tuning are determined to obtain the quantized model. To account for the model's robustness,  We consider making different choices in various components to assess the robustness of different pipelines. Specifically, it is crucial to consider whether to use adversarial training full-precision model parameters to initialize the quantization model, and whether to combine adversarial training with quantization during the training process. We aim to observe the effects of quantizing to different bit-widths on robustness under various settings. Additionally, it is important to investigate the impact of choices regarding dataset size on robustness.

\subsection{Evaluate Method}

The evaluation methods focus on the performance of the quantization model. It is generally measured using the accuracy of adversarial examples \citep{BenchmarkRobust2020}. As improving the adversarial robustness of a model may impact its accuracy, it is necessary to consider both aspects simultaneously. Therefore, we propose using adversarial accuracy as a metric for robustness (denoted as $Robustness$) while also calculating its accuracy on clean samples (denoted as $Accuracy$).

We use attack model $\mathcal{A}_{\epsilon,p}$ with budget $\epsilon$ under the $L_p$ norm to generate adversarial sample  $x_{adv}^i=\mathcal{A}_{\epsilon,p}(x^i)$.
For the upcoming evaluation of model $\mathcal{M}$, we can compute the accuracy using the formula
$Accuracy(\mathcal{M})=\frac{1}{N}\sum_{i=1}^N{\displaystyle \sI(x^i=y_i)}$, and compute the robust using the formula$\ Robustness(\mathcal{M})=\frac{1}{N}\sum_{i=1}^N{\displaystyle \sI(\mathcal{A}_{\epsilon,p}(x^i)=y_i)},$
where $\{x_i, y_i\}$ is the test set and $\displaystyle \sI(\cdot)$ is the indicator function.

In addition, due to the addition of robustness components during the quantization process, there may be additional time overhead. This could be particularly challenging for resource-constrained scenarios. Therefore, we compare the relative time overheads of different methods by theoretical analysis to assess their impact.

\section{Experiments and Insights}
In this section, we will discuss in detail the impact of different pipeline configurations on the robustness of quantized models. In Section 4.1, we will explicitly define the various components present in the pipeline. Furthermore, for each component, we will delve into a detailed discussion of the potential different choices that may exist. In Section 4.2, we will elaborate on the specific experimental design. Section 4.3 will then present and analyze the experimental results.

\subsection{Various Components in Quantization Pipeline}

{\bf FP Model Parameters.} 
In PTQ, adopting the parameters of the full-precision model is necessary for initialization \citep{li2021brecq}. 
In QAT, the model can be trained from scratch with random initialization or fine-tuned using parameters initialized from the full-precision model \citep{tang2022mixed,tang2023seam} for fast convergence. 
We aim to understand the impact of initialization parameters that have undergone adversarial training on the quantized model.

{\bf Quantization Precision.}
Before starting the quantization process, it is necessary to assess the deployment environment to determine the desired bit-width for quantization. There is a trade-off between computational resources and accuracy. Generally, lower quantization bit-widths have lower requirements for computational resources, but the accuracy will decrease \citep{esser2019lsq,li2021brecq}. However, the impact of precision on adversarial robustness has not been widely explored yet. In this part, we focus on the impact of the bit-width of quantized networks on the model's robustness. 

{\bf Training Strategy (for QAT).}
During the QAT process, the quantized network undergoes training or fine-tuning. We aim to understand the impact on the robustness of the quantized network when using the incorporating adversarial training method compared to original training. Due to the coupling of training strategies and initialization parameters, we need to consider all possible scenarios.

{\bf Other Adversarial Learning Method.} 
In addition, some robust learning methods are inherently suitable for quantized models. Approaches like the robust transfer network \citep{transferring2022} can obtain both robustness and accuracy through knowledge distillation from a well-trained robust model. We aim to determine whether this method is equally applicable to quantized models.

\subsection{Experimental Setup}
{\bf Dataset and Network Architecture.}
We aim to investigate the issues of adversarial robustness of quantization models in widely used computer vision tasks. Therefore, we have chosen the fundamental task of image classification for experimentation. The dataset used in the experiments is CIFAR-10 \citep{cifar10}, which is commonly employed for adversarial attack scenarios. The network architecture selected for the experiments is ResNet20 \citep{resnet}, a proven structure known for its effective application across various tasks. 

{\bf Quantization Methods.} 
We adopt the PWLQ method \citep{pwlq} as the PTQ approach for our experiments. PWLQ employs a segmented linear quantization technique, providing finer-grained quantization values for intervals with more data distribution. This allows better accuracy at lower bit-width. For QAT, we choose PACT \citep{pact} as the experimental method, which introduces learnable bounds of activations to filter outliers adaptively.  For a fair comparison, we set the fine-tuning training to 60 epochs, while training from scratch is 200 epochs.

{\bf Adversarial Attack and Defense.}
When discussing the robustness introduced by quantization itself, it is essential to measure performance under both white-box and black-box attack scenarios to avoid the obfuscated gradients \citep{obfuscated-gradients} in measured robustness. For white-box attacks, we choose the PGD attack method \citep{pgd}, known as the strongest first-order white-box attack method. For black-box attacks, AutoAttack \citep{autoattack} is selected, known for its outstanding performance and widespread use in robustness evaluations. In examining the impact of robustness components on quantized networks, we opt for the PGD adversarial training method \citep{pgd} to obtain robustness for the corresponding components. We measure the accuracy and robustness of the quantized model through PGD attacks.

\subsection{Results and Analysis}

\begin{wrapfigure}[15]{r}{0.6\textwidth}
\vspace{-1.0cm}
\begin{center}
\includegraphics[width=0.6\textwidth]{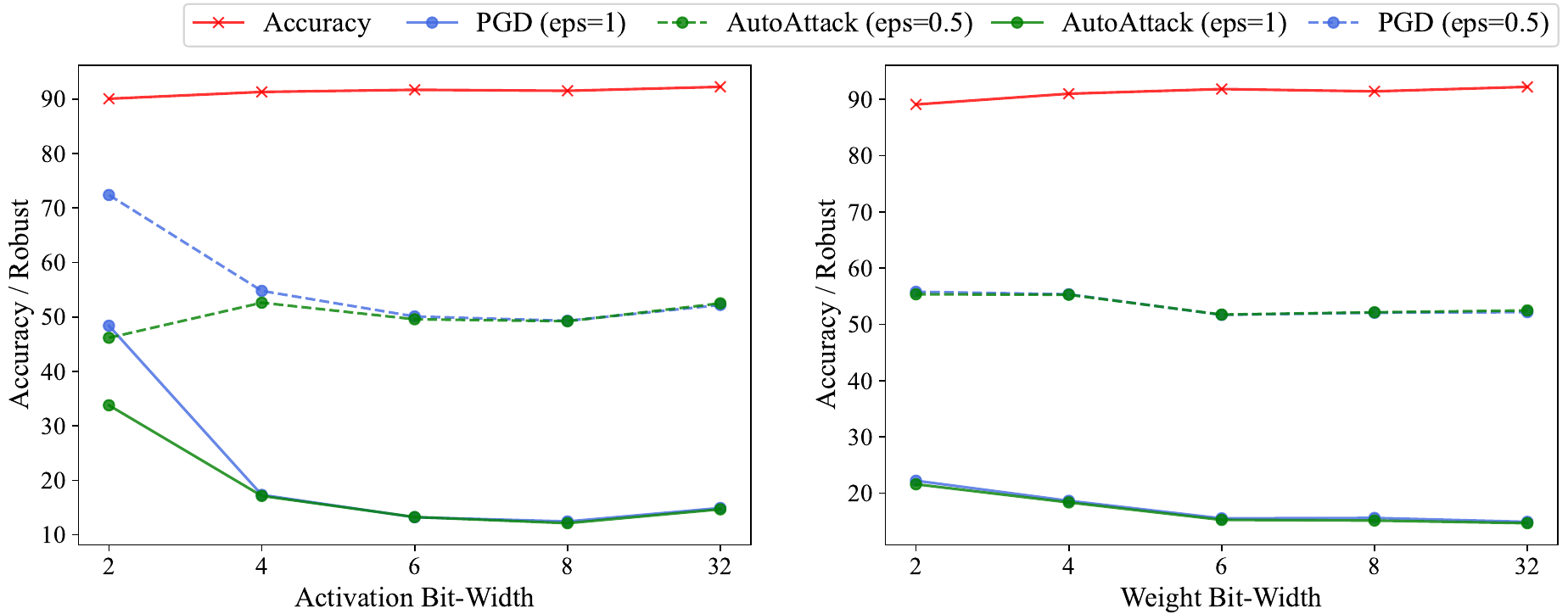}
\end{center}
\caption{Accuracy and robustness under different attacks in QAT. The left diagram represents the results of keeping weights in full precision and quantizing activations, while the right diagram represents the results of keeping activations in full precision and quantizing weights. }
\label{acc-robust}
\end{wrapfigure}

Figure \ref{acc-robust} shows the accuracy and robustness under different attacks of quantized networks obtained through random initialization in QAT. We observe that in quantized networks without any robustness components, they are unable to resist attacks with common attack settings ($\epsilon=8/255$). However, as the attack intensity decreases, both PTQ and QAT quantized networks exhibit an increase in robustness with the reduction of quantization bit-width. We also observe that activation quantization enhances robustness more noticeably compared to weight quantization. For instance, under the PGD-20 attack with $\epsilon=1/255$, quantizing activations from full precision to 2-bit results in a robustness improvement of $33.47\%$. In the same setting, weight quantization only achieves a robustness improvement of $7.31\%$.

\begin{table}[t]
\caption{The impact of FP model parameters and training strategies on accuracy and robustness. ``NC'' represents non convergence. }
\vspace{-0.3cm}
\label{init-train}
\begin{center}
\resizebox{\textwidth}{!}{
\renewcommand{\arraystretch}{0.3}
\begin{tabular}{ccccccc}
\toprule[1.2pt] 
\multicolumn{1}{c}{\bf \makecell[c]{Quantization Methods}} 
&\multicolumn{1}{c}{\bf \makecell[c]{FP Model Parameters}}
&\multicolumn{1}{c}{\bf \makecell[c]{A2W2}}  
&\multicolumn{1}{c}{\bf \makecell[c]{A4W4}}
&\multicolumn{1}{c}{\bf \makecell[c]{A6W6}} 
&\multicolumn{1}{c}{\bf \makecell[c]{A8W8}}  
&\multicolumn{1}{c}{\bf \makecell[c]{A32W32}}  
\\  \cline{1-7} \\
\multirow{3}{*}{PTQ}
    &\makecell[c]{Non-Robust Parameters}
    &{\makecell[c]{NC}}
    &{\makecell[c]{87.91 / 0.02}}
    &{\makecell[c]{91.84 / 0.00}}
    &{\makecell[c]{91.98 / 0.00}}
    &{\makecell[c]{92.15 / 0.00}}
\\  \cline{2-7} \\
    ~
    &\makecell[c]{Robust Parameters}
    &{\makecell[c]{NC}}
    &{\makecell[c]{76.16 / 47.84}}
    &{\makecell[c]{76.95 / 47.74}}
    &{\makecell[c]{76.95 / 46.58}}
    &{\makecell[c]{77.06 / 46.43}}
\\  \cline{1-7} \\
\multirow{6}{*}{\makecell[c]{QAT\\(Vanilla Training)}}
    &\makecell[c]{Random Parameters}
    &{\makecell[c]{86.99 / 0.01}}
    &{\makecell[c]{91.18 / 0.00}}
    &{\makecell[c]{91.34 / 0.00}}
    &{\makecell[c]{91.27 / 0.00}}
    &{\makecell[c]{91.71 / 0.00}}
\\  \cline{2-7} \\  
    &\makecell[c]{Non-Robust Parameters}
    &{\makecell[c]{NC}}
    &{\makecell[c]{91.66 / 0.00}}
    &{\makecell[c]{92.18 / 0.00}}
    &{\makecell[c]{92.33 / 0.00}}
    &{\makecell[c]{92.24 / 0.00}}
\\  \cline{2-7} \\
    ~
    &\makecell[c]{Robust Parameters}
    &{\makecell[c]{NC}}
    &{\makecell[c]{88.95 / 0.10}}
    &{\makecell[c]{90.28 / 0.13}}
    &{\makecell[c]{90.48 / 0.03}}
    &{\makecell[c]{90.93 / 0.00}}
\\  \cline{1-7} \\
\multirow{6}{*}{\makecell[c]{QAT\\(Adversarial Training)}}
    &\makecell[c]{Random Parameters}
    &{\makecell[c]{NC}}
    &{\makecell[c]{72.17 / 45.83}}
    &{\makecell[c]{74.04 / 46.48}}
    &{\makecell[c]{74.41 / 46.79}}
    &{\makecell[c]{76.30 / 46.83}}
\\  \cline{2-7} \\
    ~
    &\makecell[c]{Non-Robust Parameters}
    &{\makecell[c]{NC}}
    &{\makecell[c]{64.95 / 50.29}}
    &{\makecell[c]{67.51 / 37.14}}
    &{\makecell[c]{67.23 / 36.89}}
    &{\makecell[c]{70.65 / 41.25}}
\\  \cline{2-7} \\
    ~
    &\makecell[c]{Robust Parameters}
    &{\makecell[c]{NC}}
    &{\makecell[c]{76.18 / 48.48}}
    &{\makecell[c]{77.62 / 47.57}}
    &{\makecell[c]{77.41 / 47.36}}
    &{\makecell[c]{77.46 / 46.34}}
\\  \bottomrule[1.2pt]
\end{tabular}
}
\vspace{-0.5cm}
\end{center}
\end{table}

Table \ref{init-train} displays the accuracy and robustness under $\epsilon=8/255$ of quantized models obtained using different FP model parameters and selecting various training strategies. We found that models obtained through PTQ using robust parameters can still maintain the original robustness. The robustness and accuracy decrease as the quantization bit-width decreases. This may be because PTQ does not fine-tune the model but only affects its performance by introducing quantization noise. Therefore, it does not have an additional impact on robustness. For QAT, despite introducing robust initialization parameters, after fine-tuning on a dataset with clean samples, the robustness provided by the initialization diminishes to $0.1\%$. However, there is a $11.89\%$ increase in accuracy.
When adversarial training is added to the training process of QAT, it can maintain the robustness provided by the initialization. Even if random initialization is provided, QAT can still achieve robustness through adversarial training.
We also observed that when the provided initialization parameters are obtained without adversarial training, using adversarial training in QAT can lead to a significant decrease in model accuracy. For instance, in the A4W4 setting, its accuracy drops by $12.11\%$ compared to the full-precision robust model.

Incorporating adversarial training into the training strategy of QAT introduces a significant amount of additional time overhead. As each training epoch requires the generation of adversarial samples, following the classical PGD-7 adversarial training approach \citep{pgd}, each training epoch involves seven backward pass computations for adversarial samples, and a backward pass is performed to achieve gradient optimization. This results in a sevenfold increase in time overhead compared to regular training.

\begin{wraptable}{r}{0.6\textwidth}
\vspace{-1.0cm}
\caption{Accuracy and robustness on robust transfer learning with various dataset sizes.}
\label{adv-transfer}
\begin{center}
\resizebox{0.6\textwidth}{!}{
\renewcommand{\arraystretch}{0.3}
\begin{tabular}{ccccc}
\toprule[1.2pt] 
\multicolumn{1}{c}{\bf \makecell[c]{Datasets Size}} 
&\multicolumn{1}{c}{\bf \makecell[c]{A4W4}}
&\multicolumn{1}{c}{\bf \makecell[c]{A6W6}}  
&\multicolumn{1}{c}{\bf \makecell[c]{A8W8}}
&\multicolumn{1}{c}{\bf \makecell[c]{A32W32}}
\\  \cline{1-5} \\
    \makecell[c]{1\%}
    &{\makecell[c]{66.89 / 43.42}}
    &{\makecell[c]{76.45 / 47.93}}
    &{\makecell[c]{77.85 / 45.65}}
    &{\makecell[c]{77.24 / 46.86}}
\\  \cline{1-5} \\
    \makecell[c]{10\%}
    &{\makecell[c]{73.78 / 44.75}}
    &{\makecell[c]{76.69 / 48.21}}
    &{\makecell[c]{77.35 / 48.10}}
    &{\makecell[c]{77.05 / 47.51}}
\\  \cline{1-5} \\
    \makecell[c]{100\%}
    &{\makecell[c]{74.40 / 45.10}}
    &{\makecell[c]{76.41 / 48.50}}
    &{\makecell[c]{76.72 / 48.16}}
    &{\makecell[c]{76.71 / 47.87}}
\\  \bottomrule[1.2pt]
\end{tabular}
}
\vspace{-0.5cm}
\end{center}
\end{wraptable}

Table \ref{adv-transfer} represents the results when QAT with the robust transfer learning method proposed by \cite{transferring2022} from robust initialization parameters, and other settings will be shown in Appendix B. We analyzed the effect of robust transfer learning using a partial dataset. It can be observed that this method still performs well when transferred to quantized models. At higher quantization bit-widths, it can even surpass the robustness of full-precision models. For example, in the A6W6 quantization setting, its robustness increases by $2.07\%$ compared to the full-precision adversarially trained model. However, as the quantization bit-width decreases, both accuracy and robustness experience a slight decline, possibly due to a reduction in model capacity with decreasing bit-width. Additionally, even when quantizing models using a partial dataset, good accuracy and robustness can still be achieved. It is worth noting that, due to the absence of the time required to compute adversarial samples, this method is more efficient in terms of quantizing time compared to the QAT approach combined with adversarial training.

\section{Conclusion}
In this paper, we first introduce the evaluation metrics for adversarial robustness in quantized networks. By studying the impact of quantization itself on robustness and the effects of adding robustness components, we aim to provide valuable insights for the design and evaluation of robust quantization pipelines. We hope that this work can serve as a useful step in improving robustness in quantized networks.

\subsubsection*{Acknowledgments}
This work was supported in part by the National Key Research and Development Project of China (Grant No. 2023YFF0905502), Shenzhen Science and Technology Program (Grant No. RCYX20200714114523079 and JCYJ20220818101014030).

\bibliography{iclr2024_conference}
\bibliographystyle{iclr2024_conference}
\newpage
\appendix
\section{Attack Strength and Quantization Settings}
\textbf{Adversarial Attack Intensity.} 
The adversarial sample construction in adversarial attacks can be denoted as $x_{adv}=x+argmin_\delta\{||\delta||:f(x+\delta) \neq f(x), \delta \leq \epsilon\}$, where $\epsilon$ is a small quantity representing the magnitude of a slight perturbation applied to the input sample. In PGD attacks conducted on the CIFAR-10 dataset, it is common to set $\epsilon \leq 8/255$, so we define it as a ``normal attack". When $\epsilon \leq 2/255$, full-precision model without adversarial training may exhibit a certain level of robustness which is referred to as a ``weaker attack".

\textbf{Quantization Settings.} 
Quantization settings include PTQ and QAT. For the QAT method, it requires either Fine-Tuning or Training. Specifically, the training strategy using the original QAT is referred to as QAT (Vanilla Training). When generating adversarial samples in the quantized model and employing Adversarial Training (\citealp{pgd}), the training strategy is termed QAT (Adversarial Training). The approach of using the quantized model as a student model for adversarial transfer learning (\citealp{transferring2022}) is referred to as QAT (Adversarial Transfer Learning).

\section{More Details about Transfer Adversarial Learning}

\begin{wraptable}{r}{0.65\textwidth}
\vspace{-0.75cm}
\caption{Accuracy of Transfer Adversarial Learning under Different Initialization. The Accuracy of the Teacher Model (Full-Precision Model) is $77.06\%$.}
\vspace{-0.25cm}
\label{}
\begin{center}
\resizebox{0.65\textwidth}{!}{
\renewcommand{\arraystretch}{0.3}
\begin{tabular}{cccccc}
\toprule[1.2pt] 
\multicolumn{1}{c}{\bf \makecell[c]{Initializing Parameters}}
&\multicolumn{1}{c}{\bf \makecell[c]{A3W3}}  
&\multicolumn{1}{c}{\bf \makecell[c]{A4W4}}
&\multicolumn{1}{c}{\bf \makecell[c]{A6W6}} 
&\multicolumn{1}{c}{\bf \makecell[c]{A8W8}}  
&\multicolumn{1}{c}{\bf \makecell[c]{A32W32}}  
\\  \cline{1-6} \\
    \makecell[c]{Random Parameters}
    &{\makecell[c]{72.93}}
    &{\makecell[c]{74.21}}
    &{\makecell[c]{74.62}}
    &{\makecell[c]{74.26}}
    &{\makecell[c]{72.18}}
\\  \cline{1-6} \\
    \makecell[c]{Non-Robust Parameters}
    &{\makecell[c]{67.83}}
    &{\makecell[c]{69.98}}
    &{\makecell[c]{69.90}}
    &{\makecell[c]{70.23}}
    &{\makecell[c]{71.55}}
\\  \cline{1-6} \\
    \makecell[c]{Robust Parameters}
    &{\makecell[c]{NC}}
    &{\makecell[c]{74.40}}
    &{\makecell[c]{76.41}}
    &{\makecell[c]{76.72}}
    &{\makecell[c]{76.71}}
\\  \bottomrule[1.2pt]
\end{tabular}
}
\end{center}

\vspace{0.2cm}
\caption{Robustness of Transfer Adversarial Learning under Different Initialization. The Robustness of the Teacher Model (Full-Precision Model) is $46.43\%$.}
\vspace{-0.25cm}
\label{}
\begin{center}
\resizebox{0.65\textwidth}{!}{
\renewcommand{\arraystretch}{0.3}
\begin{tabular}{cccccc}
\toprule[1.2pt] 
\multicolumn{1}{c}{\bf \makecell[c]{Initializing Parameters}}
&\multicolumn{1}{c}{\bf \makecell[c]{A3W3}}  
&\multicolumn{1}{c}{\bf \makecell[c]{A4W4}}
&\multicolumn{1}{c}{\bf \makecell[c]{A6W6}} 
&\multicolumn{1}{c}{\bf \makecell[c]{A8W8}}  
&\multicolumn{1}{c}{\bf \makecell[c]{A32W32}}  
\\  \cline{1-6} \\
    \makecell[c]{Random Parameters}
    &{\makecell[c]{29.93}}
    &{\makecell[c]{34.58}}
    &{\makecell[c]{35.66}}
    &{\makecell[c]{34.75}}
    &{\makecell[c]{40.66}}
\\  \cline{1-6} \\
    \makecell[c]{Non-Robust Parameters}
    &{\makecell[c]{16.97}}
    &{\makecell[c]{19.40}}
    &{\makecell[c]{18.71}}
    &{\makecell[c]{18.52}}
    &{\makecell[c]{31.64}}
\\  \cline{1-6} \\
    \makecell[c]{Robust Parameters}
    &{\makecell[c]{NC}}
    &{\makecell[c]{45.10}}
    &{\makecell[c]{48.50}}
    &{\makecell[c]{48.16}}
    &{\makecell[c]{47.87}}
\\  \bottomrule[1.2pt]
\end{tabular}
}

\vspace{-0.2cm}
\end{center}
\end{wraptable}

For Transfer Adversarial Learning, the first step involves training a robust full-precision model as the teacher model. Subsequently, based on the chosen quantization bit-width, a quantized model will serve as the student model, employing the loss function proposed by \cite{transferring2022} for knowledge distillation, allowing simultaneous learning of both accuracy and robustness.

In this method, the initialization parameters of the student model can also be chosen. We set them to be initialized from an adversarial trained full-precision model (robust initializing parameters), a vanilla trained full-precision model (non-robust initializing parameters), and a scratch-trained model.  Simultaneously, we specify the teacher model to be an adversarial trained full-precision model for experimentation.

Table 4 presents the accuracy obtained under different initialization parameters and quantization bit-widths, while Table 5 displays the robustness. 
In robust and non-robust initialization methods, accuracy consistently demonstrates a decrease with the reduction of quantization bit-widths.
For robustness, it can be observed that initializing the student model with a robust full-precision model yields the best robustness. In fact, in certain quantization bit-width, it even surpasses the robustness of the teacher model. For instance, the A8W8 configuration exhibits a $1.73\%$ improvement in robustness compared to the full-precision robust model. However, in this setup, robustness experiences a slight decline with decreasing quantization bit-width. 

The method of initializing with non-robust parameters or training from scratch exhibits a significant decline in robustness performance on the quantized network compared to the full-precision network. For instance, in the case of non-robust initialization, using an A3W3 quantized network results in a $14.67\%$ decrease in robustness compared to the full-precision network. This indicates the crucial role of model initialization in robustness within the context of Transfer Adversarial Learning.

\end{document}